\documentclass[10pt,letterpaper]{article}

\usepackage{cogsci}
\usepackage{graphicx}

\cogscifinalcopy %%% Uncomment this line for the final submission

\usepackage[
  backend=biber,
  style=apa,
  natbib=true,
  annotation=false,
]{biblatex}
\usepackage{csquotes}
\usepackage{float} 

\title{Are you with me?  A Framework for Detecting Mental Model Discrepancies in Task-Based Team Dialogues}

\author[1]{\mbox{Katharine Kowalyshyn (katharine.kowalyshyn@tufts.edu)}}
\author[1]{\mbox{Matthias Scheutz}}
\affil[1]{Department of Computer Science, Tufts University}

\begin{document}

\maketitle

\begin{abstract}
  Humans typically use natural language to update teammates on task
  states.  Since not all updates are communicated, discrepancies arise
  between the team members' mental models that negatively affect
  overall team performance.  How can we categorize such discrepancies?
  Do misalignments detected in team dialogue predict future mental
  model misalignments?  Traditional shared mental model (SMM)
  assessment methods rely on retrospective expert coding that cannot
  capture real-time coordination dynamics. We propose a framework to
  identify and categorize four types of mental model discrepancies:
  {\em unsupported beliefs}, {\em false beliefs}, {\em belief
    contradictions}, and {\em omissions}, all of which can naturally
  emerge in team dialogues. Using dialogues from twenty dyad teams
  performing collaborative object identification tasks across four
  sequential levels, we demonstrate that these discrepancy patterns
  contain predictive signals. Averaging historical discrepancy counts
  achieves meaningful prediction accuracy using uniform weighting as an exploratory baseline, with
  differential predictability across discrepancy types.

\end{abstract}

\section{Introduction}

Effective teamwork depends fundamentally on coordination: the ability
of multiple agents to work together toward shared goals while managing
interdependencies in their actions. In many real-world scenarios, from
surgical teams to search and rescue teams, team members must
coordinate their activities through natural language dialogues,
especially when not all team members are co-present.  These updates
ensure that each team member has correct representations of the current
task state, the various activities and subgoals of other team members,
and other task-relevant factors in their {\em mental model} (MM). When
team members successfully maintain overlapping or compatible mental
models, they achieve a \textit{Shared Mental Model} (SMM)
\parencite{mathieu2000influence}.

When important task updates are not communicated, the individual team
members' mental models can diverge, i.e., teammates that are privy to
the task changes will update their mental models while those not able
to obtain the information will still have false beliefs about the
previous task state in their mental model.  If not resolved through
communicated updates, these inconsistencies can lead to coordination
breakdowns, execution errors, and complete task failure. Understanding
how and when these breakdowns occur is, therefore, critical for
improving team performance \parencite{hawkins}.

In this paper, we present the first systematic categorization of SMM
discrepancy types in naturalistic team coordination tasks. Through
manual annotation of team communication and task execution data from a
collaborative task environment, we identify distinct categories of
mental model misalignment and discuss their differential impacts on
team success. Our contribution provides a theoretical framework for
categorizing different forms of SMM breakdown and tracking how these
discrepancies evolve over time.

\section{Motivation and Background}

The concept of mental models in team coordination is not new. It has
long been recognized that individuals construct internal
representations of external systems to reason about them and predict
their behavior \parencite{Johnson-Laird_1983}. In team settings, these
mental models extend beyond task knowledge to include representations
of teammates' knowledge, goals, and capabilities.  It is this shared
understanding---the shared mental model---that enables team members to
coordinate implicitly, reducing the communication overhead required
for task execution and allowing teams to operate efficiently even
under time pressure or communication constraints.  As such, SMMs
provide a fundamental understanding of a team's coherence, task
progression, and overall performance. The quality of an SMM directly
impacts team outcomes: teams with more accurate and aligned mental
models demonstrate superior coordination, faster task completion, and
fewer errors \parencite{mathieu2000influence}.  When an SMM is
``inconsistent'', i.e., when individual mental models are mutually
inconsistent and fail to align or when critical information is not
shared, team performance can and will suffer measurably \parencite{mathieu2000influence}.  

\begin{figure}
    \centering
    \includegraphics[width=0.9\linewidth]{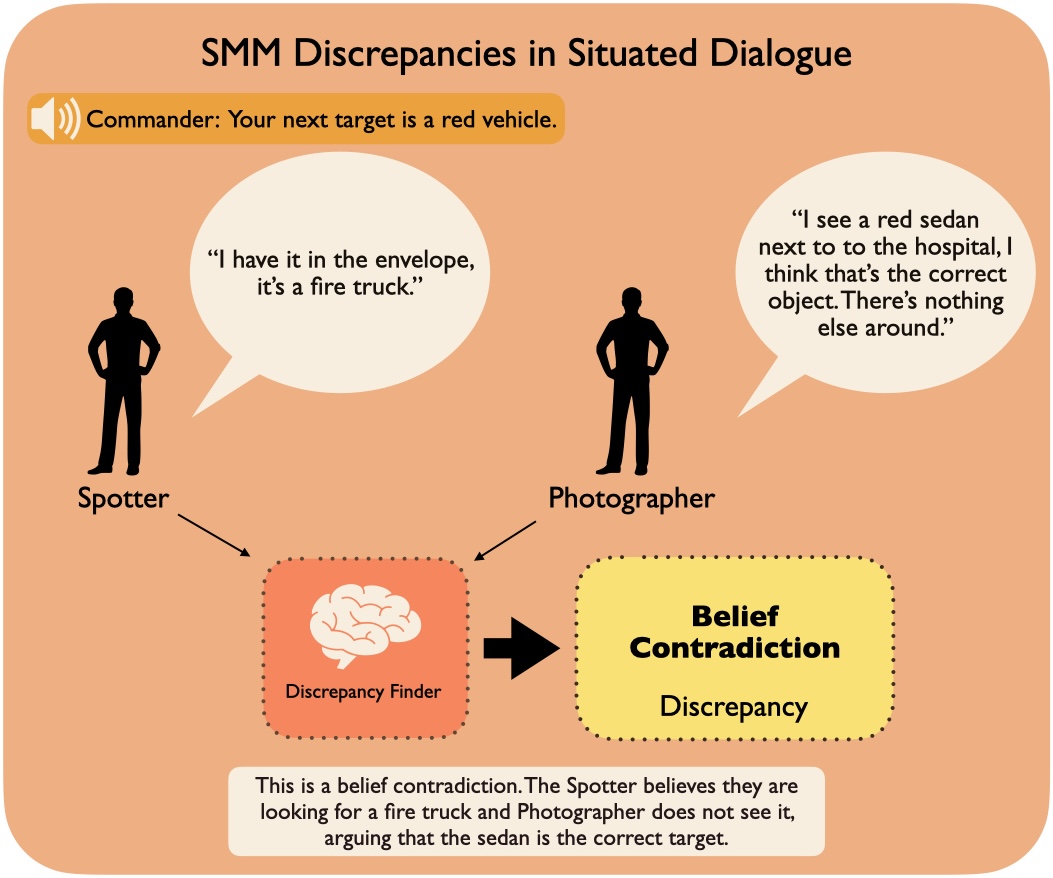}
    \caption{Example dialogue snippet from our experiments showing how
      an agent may identify a belief contradiction between the Spotter
      agent and the Photographer agent regarding object location (see
      Methodology for details on the task).}
\end{figure}

Previously, SMMs have been used as a metric post-hoc to analyze team
performance and success
\parencite{mathieu2000influence,scheutz2017framework}, with
researchers developing various methods for assessing mental model
similarity and quality. These approaches have revealed important
insights about the relationship between SMM accuracy and team outcomes
across diverse domains. Still, the prevalent challenge and limitation
of SMM use remains online team analysis. Detecting when individual
mental models diverge, creating discrepancies in the shared mental
model, is difficult with traditional methods, which often rely on
retrospective scoring or expert annotation of recorded team
interactions. These approaches cannot capture the dynamic, real-time
nature of coordination breakdowns as they occur during task
execution. Moreover, existing work has not systematically examined
what types of discrepancies exist or how different forms of mental
model misalignment impact performance differently. While we know that
SMM quality matters for team success, we lack a detailed understanding
of the specific patterns of mental model divergence that can lead to
coordination failures.

This gap in understanding motivates our research question: Can we identify and categorize specific types of SMM discrepancies and predict their patterns? We hypothesize that not all mental model
misalignments are equally problematic. Some types of discrepancies may exhibit more stable temporal patterns than others, and different categories of misalignment may follow distinct trajectories over time. Understanding how particular discrepancy patterns evolve and persist would enable
more effective team monitoring and intervention strategies, allowing
us to focus attention on the most critical forms of mental model
divergence.

\section{Related Work}

\subsection{Mental Models in Team Coordination}

Empirical work has established strong relationships between SMM
quality and team performance across diverse domains. Mathieu et
al. \parencite{mathieu2000influence} demonstrated that teams with more
similar mental models exhibit superior coordination and task outcomes,
particularly under time pressure or ambiguous conditions. This
foundational work revealed that SMM similarity matters most for
taskwork mental models (understanding of the task itself) and teamwork
mental models (understanding of roles and interaction
patterns). Subsequent research has extended these findings to dynamic
environments, showing that SMMs must be continuously updated as task
conditions change \parencite{scheutz2017framework}. Recent work has further shown that shared goals induce structured division of labor in communication \parencite{hawkins}, that teams must coordinate on shared procedural abstractions through dialogue \parencite{mccarthy2021learningcommunicatesharedprocedural}, and that communicators select sparse but relevant information during cooperation \parencite{Jiang_Jiang_Sadaghdar_Limb_Gao_2025}, all of which bear directly on when and why omissions arise in asymmetric-information tasks.

Despite this extensive body of work, most SMM research has relied on
post-hoc assessment methods, including structured interviews,
questionnaire-based similarity ratings, and retrospective expert
coding of team interactions \parencite{mohammed2000measurement, dechurch2010measuring}. 
While these approaches have yielded
important theoretical insights, they provide global assessments of
mental model similarity without distinguishing between different types
of misalignment. Furthermore, existing work has not systematically
categorized the types of discrepancies that emerge when individual
mental models diverge, nor examined how different discrepancy patterns evolve over sequential task episodes.

\subsection{Tracking Mental States in Natural Dialogue}

Understanding how team members communicate their mental states and
track their teammates' understanding through natural dialogue has been
a key focus in cognitive science and psycholinguistics especially in
regard to SMMs. Research has shown that humans are remarkably adept at
inferring others' mental states from conversational cues, including
explicit statements, implicit references, questions, and even silence
\parencite{kosinski2024evaluating}. This {\em Theory of Mind} (ToM)
reasoning enables team members to detect potential misalignments and
repair them through targeted communication.

Prior work on common ground in dialogue \parencite{clark1991grounding}
has examined how conversational partners establish and maintain mutual
understanding. However, this work has not systematically categorized
the types of failures that occur when common ground breaks down in
team coordination contexts. There is a distinction between the number
of discrepancies and what specifically those discrepancies consist of
which has been a topic of research \parencite{lim2006team}. Work on task-based dialogue has revealed, for
example, that team members use specific linguistic markers to signal
their current understanding and probe their teammates' mental
states. These include explicit status updates (``I've completed
section A''), implicit assumptions (``Now that we're done with the
first part...''), and checking questions ("Do you have the materials
ready?") \parencite{mathieu2000influence}. By
analyzing these communicative patterns, researchers can infer when
individual mental models are aligned versus divergent.

However, existing dialogue analysis in team settings has focused
on overall communication patterns or task outcomes rather than
systematically categorizing mental model discrepancies \parencite{MARLOW2018145}. Traditional
SMM assessment methods including similarity metrics comparing
individual mental models and expert ratings of team communication
effectively measure overall SMM quality but do not differentiate
between the nature of misalignments. Specifically, these approaches
cannot distinguish whether teams hold contradictory beliefs about task
state, fail to share critical information, misinterpret their
teammates' actions, or maintain unsupported assumptions \parencite{MARLOW2018145}.  
Our work addresses this gap by developing a systematic framework for
categorizing four types of mental model discrepancies observable in
natural team dialogue: {\em unsupported beliefs}, {\em false beliefs},
{\em belief contradictions}, and {\em omissions}. By examining how
these discrepancy patterns evolve across sequential coordination
episodes, we demonstrate that different types of misalignment exhibit differential temporal predictability and stability.

\begin{figure*}[t]
    \centering
    \includegraphics[width=0.48\textwidth]{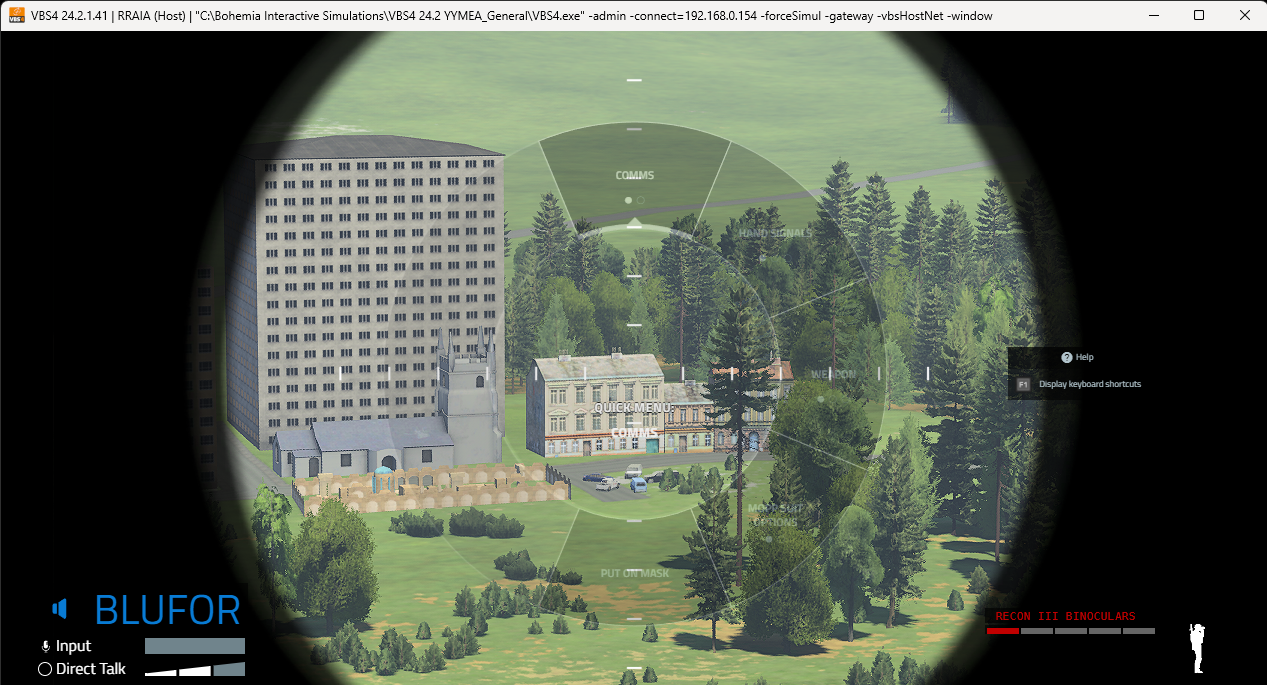}
    \hfill
    \includegraphics[width=0.48\textwidth]{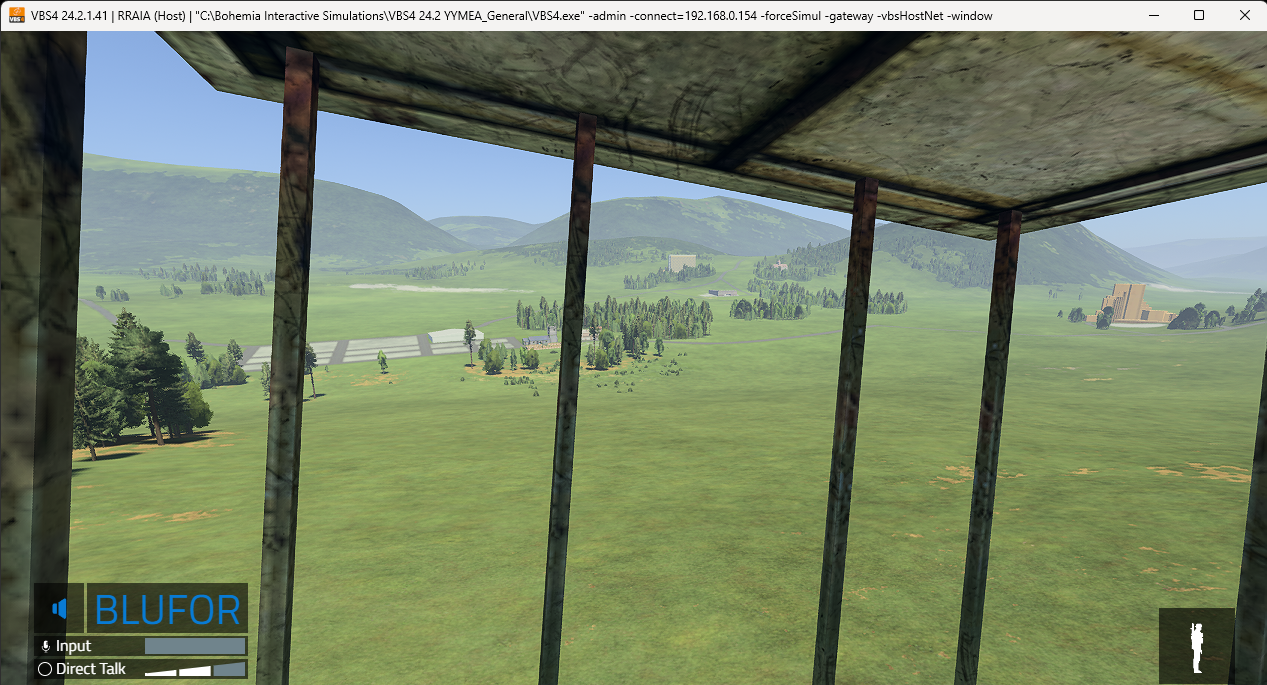}
    \caption{Left: Binoculars view of experiment environment from
      \textit{Spotter} point of view. Right: Participant view from
      tower from \textit{Photographer} point of view. Both
      participants stand on the same tower with differing ability to
      view the environment.}
\end{figure*}

\section{Methodology}
\label{sec:method}

\subsection{Experimental Setup \& Dataset}

We developed an experimental setting in which dyad teams had to
perform a collaborative object identification task in a video game
environment with asymmetric information.  The task was designed to
have four difficulty levels each lasting for eight minutes, with each
level consisting of a different environmental setting in the order:
{\em Local Law Enforcement}, {\em Mountain Search and Rescue}, {\em
  Beach Patrol}, and {\em Poaching}. We refer to these as Levels 1–4 respectively throughout the paper. Each level lasted eight minutes and was designed to be approximately equal in difficulty, allowing us to treat them as comparable episodes for discrepancy tracking. Each level had the same number of targets and we had two foils, fog and white noise, which were each used in levels 1\&3 and 2\&4 respectively. Such distractors were set to average at 325 seconds into each level, with each distractor
differing in exact timing but the same across all participants.

As seen in Figure 2, both team members occupied the same tower vantage
point but had asymmetric information access: the \textit{Spotter}
received detailed written instructions describing target objects but
had limited visual range, while the \textit{Photographer} possessed
enhanced binoculars for distant viewing but relied entirely on verbal
cues from the Spotter to locate targets. This asymmetric structure was
chosen to encourage frequent dialogue-based coordination despite
frequent shared physical positioning. Teams communicated via an audio
channel while attempting to identify and photograph specified objects
in the environment.

All communications from all teams were transcribed with the help of an
LLM and checked for correctness by human annotators. The
transcriptions were subsequently fed into another previously developed
LLM model for detecting the four types of discrepancies which
performed mental model updates at the level of human annotators (this
work is currently under review).  The resultant dataset was then used
to explore the idea of predicting team performance via our discrepancy
categorization framework below. All analyses reported in this paper are exploratory; no hypotheses were preregistered.

\subsection{Discrepancy Categorization Framework}

As mentioned before, we isolated four discrepancy patterns that may
predict future coordination breakdowns:

\begin{itemize}
    \item \textbf{Belief Contradiction:} one team member believes something
and another believes its negation \\ {\em Example:} Team Leader believes ``the building's east wing will collapse in 30 minutes'' while Safety Officer believes ``the east wing is stable for 2 hours''
    \item \textbf{Omission:} a clear discrepancy where one
team member's mental model omits another team member's belief. In practice, omissions are inferred from the asymmetry between what a team member's role implies they should know and what their communicated mental model reflects. \\
{\em Example:} Hazmat specialist knows ``We need special gear beyond 50 feet from spill'' but the team leader's mental model lacks this critical safety information
    \item \textbf{Unsupported Beliefs:} beliefs that
a team member identifies but are not contradicted nor
supported in the ground truth or by another team member's mental model. Unsupported beliefs are distinguished from false beliefs in that they cannot be verified against ground truth. No authoritative information exists to confirm or contradict them. False beliefs, by contrast, are explicitly contradicted by established ground truth.\\
{\em Example:} Search team member believes ``I heard tapping sounds from the northwest sector'' with no corroborating evidence from beacon data or other team members
    \item \textbf{False Beliefs:} incorrect annotations according to an established ground truth\\
    {\em Example:} Navigator believes ``GPS places missing hikers 3 miles north'' when ground truth shows they are actually 3 miles south
\end{itemize}

Each dialogue move was annotated with updates to each team member's
mental model and the four discrepancies definitions were employed to
determine which of them were present.  Specifically, we focused on
discrepancies regarding beliefs, goals, and commitments among the
team members to keep the mental models simple enough for comparison
while still capturing the main levers of team coordination (even
though other representations would be possible as well).

\subsection{Predicting Discrepancies in ToM Tasks}

To predict the number of discrepancies in a shared mental model from a
team's previous discrepancy patterns, we used a weighted sum of the
counts in each discrepancy category such that the number of
discrepancies $d$ at each target level $L_i$ is

\begin{equation}
    L_{\text{target}}(d) = \sum_{i \neq \text{target}} w_i L_i(d), \quad \sum w_i = 1
\end{equation} 
where $L_i$ indicates the level of the task at hand and $w_i$
represents the weight assigned to level $i$ in the prediction.

Framing this problem as a weighting of the discrepancy types allows
for increased applicability to a variety of tasks, and the optimal
weighting of discrepancy types depends on several task-specific
factors. First, the immediacy and reversibility of action shape risk
profiles: belief contradictions are most critical when immediate
synchronized decisions are needed (as they prevent or impact
coordinated action), false beliefs become most dangerous during
ongoing action (as they drive confident misdirection), and omissions
are most harmful when consequences are irreversible. Second, the
availability of ground truth constrains which discrepancies can be
detected. Without authoritative information or a ground truth, false
beliefs cannot be distinguished from correct beliefs, limiting
detection to contradictions, omissions, and unsupported
beliefs. Third, team coordination requirements matter: contradictions
are maximally disruptive when synchrony is required, while omissions
may be tolerable under distributed cognition where knowledge can be
shared on demand. These principles suggest that any automated mental
model discrepancy detection system should employ adaptive weighting
schemes that account for task phase, time pressure, consequence
severity, and information availability.

\begin{figure*}[t]
    \centering
    \includegraphics[width=0.85\textwidth]{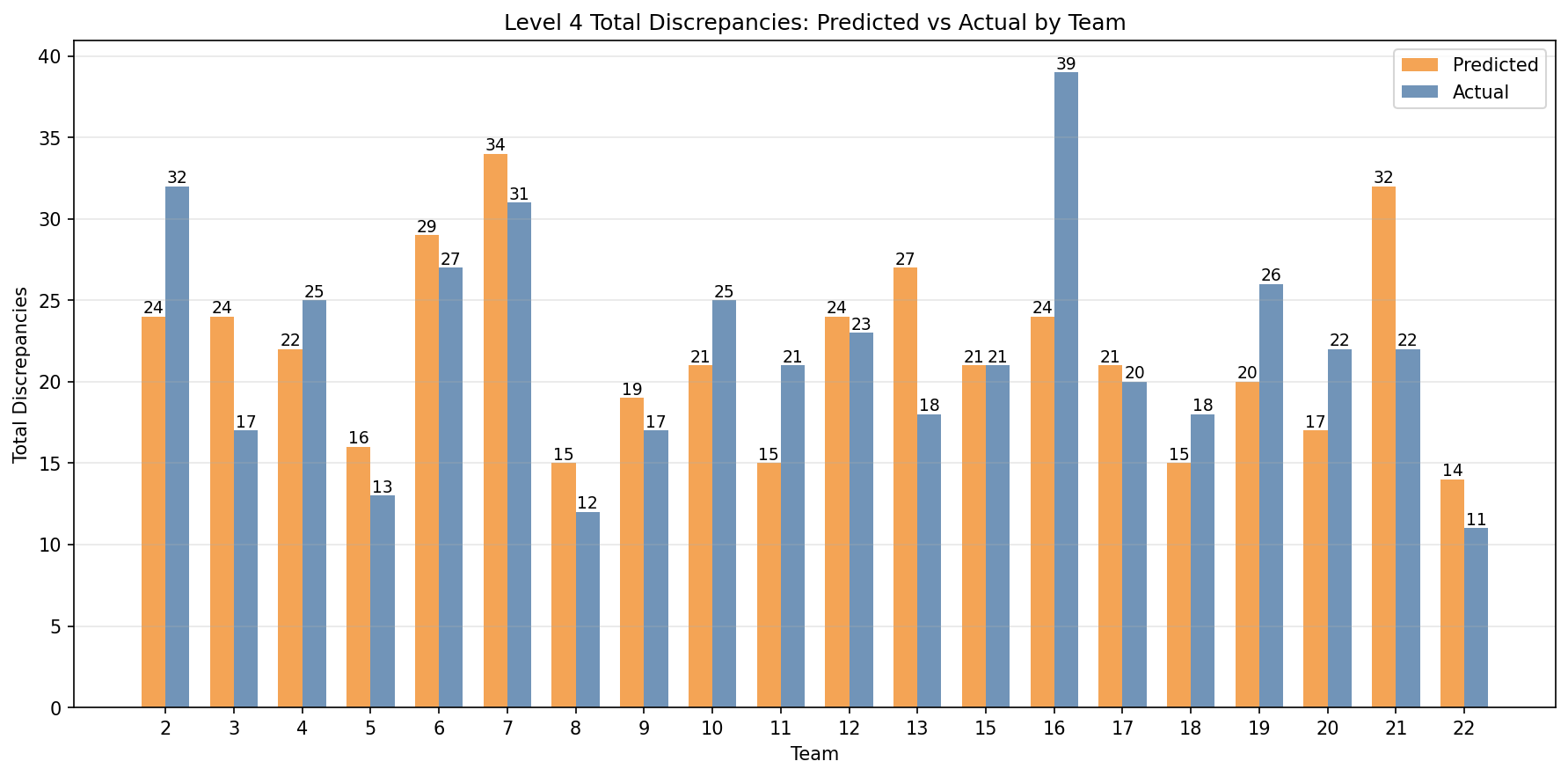}
    \caption{Level 4 total discrepancies: predicted versus actual
      counts across all teams. The baseline uniform averaging model
      shows variable but meaningful prediction accuracy, with most
      teams falling within 5 discrepancies of the predicted
      value. Teams 2, 13, and 16 show the largest prediction errors,
      suggesting these teams exhibited coordination patterns that
      deviated substantially from their earlier trajectories.}
    \label{fig:prediction-accuracy}
\end{figure*}

\subsubsection{Baseline Model and Extensions}

As a baseline model, we employ uniform weighting where $w_i =
\frac{1}{n}$ and $n$ is the number of predictor levels. This uniform
approach assumes that all prior levels contribute equally to
predicting future performance, providing a straightforward baseline
against which more sophisticated weighting schemes can be
compared. In this work, we implement only the uniform baseline $(w_i = \frac{1}{n})$. The recency-weighted and type-specific extensions described above are theoretical proposals left for future work; no alternative weighting schemes were estimated from data or empirically compared in this study.

\section{Results}

The analysis of our experimental results revealed systematic patterns
of how mental model misalignments emerge, persist, and relate to team
coordination quality, as well as served as a test for our discrepancy
framework.

\begin{figure}[H]
    \centering
    \includegraphics[width=0.9\linewidth]{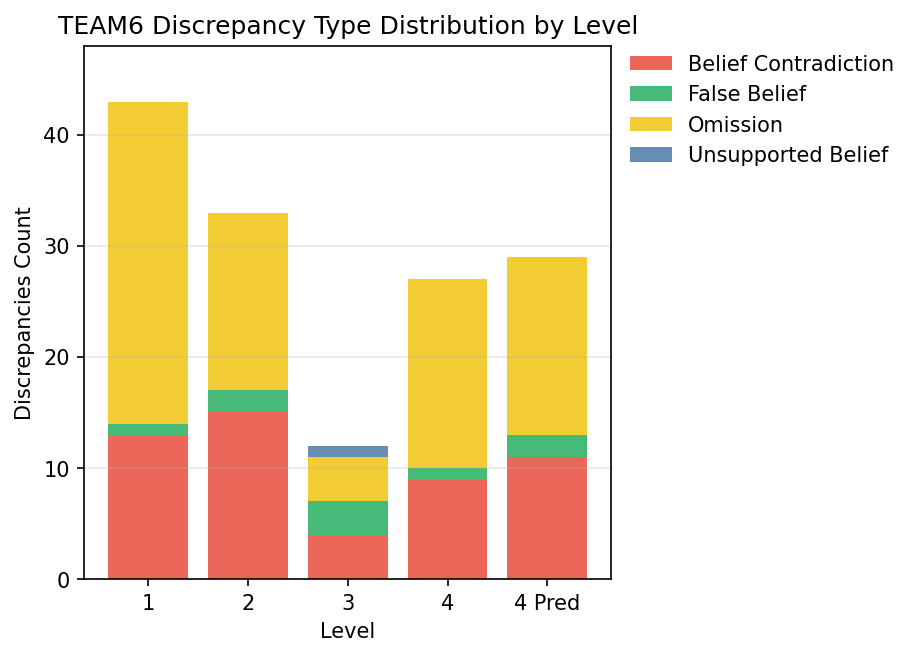}
    \caption{Team 6's discrepancy type distribution across levels. This team exhibited high initial discrepancies, a dramatic drop in Level 3, but then reverted to elevated discrepancies in Level 4. Omissions (yellow) consistently dominated, while Belief Contradictions (blue) showed the most volatility across levels.}
    \label{fig:team6-trajectory}
\end{figure}

\subsection{Discrepancy Distribution and Patterns}

Across all teams and levels, we identified a total of 1,447
discrepancies, with substantial variation both between teams (range:
44-176 total discrepancies across all four levels) and within
individual team trajectories. Figure~\ref{fig:prediction-accuracy}
presents Level 4 prediction accuracy using our baseline uniform
averaging model, comparing predicted discrepancy counts (based on
Levels 1-3) against actual observed counts.

The prediction model captured general team coordination trends but
showed varying accuracy across teams. Our pairwise correlation for all twenty teams was 0.56 with a p-value of 0.01. This correlation should be interpreted cautiously: teams with consistently high or low discrepancy rates across all levels would produce this result even without genuine predictive signal, as Level 4 counts are likely autocorrelated with earlier levels. The correlation may therefore reflect stability in team-specific baseline rates rather than evidence that discrepancy patterns mechanistically predict future coordination outcomes. Several teams (e.g., Teams 12, 15, 17) demonstrated highly predictable patterns with minimal error
between predicted and actual discrepancy counts, while others (Teams
13, 16) showed larger deviations, indicating coordination dynamics
that changed substantially in the final level. 

\subsection{Case Study: Team 6}

To illustrate how discrepancy patterns evolve across levels, we
examine Team 6 in detail. Figure~\ref{fig:team6-trajectory} shows this
team's discrepancy distribution across all four levels and the
prediction for Level 4.

Team 6 maintained persistently high discrepancy counts overall, with
the anomalous drop in Level 3 representing temporary coordination
success that failed to stabilize. The prediction model reasonably
approximated the Level 4 outcome, suggesting the team's baseline
coordination challenges persisted despite the Level 3
improvement. Omissions remained the dominant discrepancy type
throughout, while belief contradictions decreased from Level 1 to
Level 4, indicating some resolution of interpretive disagreements even
as information-sharing gaps persisted.

\subsection{Discrepancy Types and Coordination Quality}

\begin{figure}[H]
    \centering
    \includegraphics[width=0.9\linewidth]{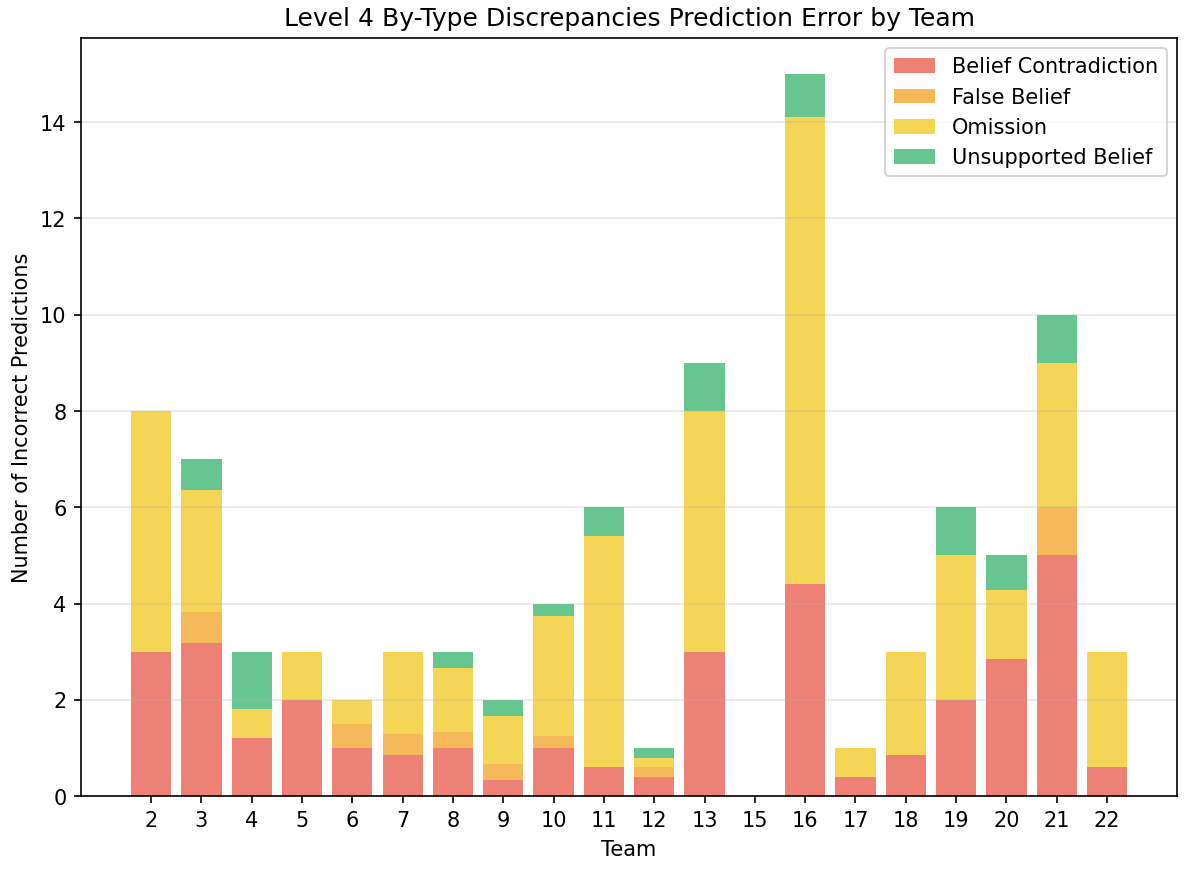}
    \caption{Level 4 prediction errors by discrepancy type across all teams. Omissions (yellow) constitute the largest source of prediction error in absolute terms, reflecting their dominance in overall discrepancy counts. Belief Contradictions (red) show moderate error magnitudes. False Beliefs and Unsupported Beliefs (green and orange) remain rare and contribute minimally to prediction error.}
    \label{fig:discrepancy-types-level4}
\end{figure}

Breaking down discrepancies by type revealed differential predictability across discrepancy categories. Figure~\ref{fig:discrepancy-types-level4} presents the prediction error analysis by discrepancy type across teams for Level
4.

Omissions showed the most consistent patterns across teams and
constituted the largest component of both total discrepancies and
prediction errors. The absolute magnitude of omission errors reflects
their frequency rather than unpredictability. Belief contradictions
demonstrated greater relative volatility, with some teams showing
substantial error while others were predicted accurately. False
beliefs and unsupported beliefs remained sufficiently rare that they
contributed negligibly to overall prediction error, appearing in fewer
than 20\% of team-level observations and typically numbering 0-2 per
level when present. The differential predictability across discrepancy types suggests that
information-sharing patterns (captured by Omissions) stabilize more
quickly than interpretive alignment (captured by Belief
Contradictions). Teams that maintained high omission rates in early
levels may have struggled with comprehensive information
sharing, while tendency toward belief contradictions showed
more variability.

\section{Discussion \& Conclusion}

Even a simple baseline prediction model using
uniform averaging of historical discrepancies achieved reasonable
accuracy for most teams, suggesting that early-stage coordination
patterns establish trajectories that persist across multiple task
instances. 

\subsection{Validation of LLM-Based Mental Model Discrepancy Detection}

To validate our LLM-based approach for detecting shared mental model discrepancies, we developed an objective scoring metric based on target identification performance. Teams were scored on their ability to visually confirm three targets: (1) an open field with kangaroos and a poacher (Hard, 6 points), (2) a building helipad with helicopter and suited individual (Easy, 6 points), and (3) a water tower with white cargo vehicle (Easy, 7 points), for a maximum of 19 points. Points were awarded only for elements teams explicitly confirmed seeing, not merely hearing about in briefings. As shown in Table 1, Team 8 achieved 8/19 points (42.1\%) compared to Team 16's 5/19 points (26.3\%), a difference of 3 points. Notably, neither team successfully identified Target 1, suggesting genuine task difficulty. The performance gap emerged primarily in Targets 2 and 3, where Team 16's coordination failures were most evident. Team 16 exhibited multiple instances of shared mental model breakdown, including the photographer's statement that ``It's giving us different stuff'' despite both teams receiving identical briefings, and confusion about whether they were observing the same water tower. In contrast, Team 8 maintained consistent shared situation awareness. These objective performance differences strongly correlate with our LLM's assessment that Team 16 demonstrated significantly more SMM discrepancies, providing empirical validation that communication-based mental model misalignment manifests as measurable decrements in collaborative task performance.

\begin{table}[t]
\centering
\label{tab:target_scores}
\small
\begin{tabular}{lccc}
\hline
\textbf{Target} & \textbf{Max} & \textbf{Team 8} & \textbf{Team 16} \\
\hline
Target 1 (Hard) & 6 & 0 (0\%) & 0 (0\%) \\
Target 2 (Easy) & 6 & 3 (50\%) & 1 (16.7\%) \\
Target 3 (Easy) & 7 & 5 (71.4\%) & 4 (57.1\%) \\
\hline
\textbf{Total} & \textbf{19} & \textbf{8 (42.1\%)} & \textbf{5 (26.3\%)} \\
\hline
\end{tabular}
\caption{Target Identification Performance by Teams 8 and 16 via our objective scoring metric.}
\end{table}
\subsection{Theoretical Implications}

These findings refine SMM theory by revealing that not all mental model misalignments follow equivalent temporal patterns. Omissions (failures to share critical information)
dominated overall discrepancy counts and showed the most stable
patterns across teams and levels. This stability suggests that
information-sharing behaviors, once established, prove resistant to
change even as teams gain experience working together.

Unlike prior methods yielding global similarity scores, our typed, temporally-resolved signal enables targeted intervention, distinguishing whether to address information-sharing gaps (omissions) versus interpretive misalignment (belief contradictions).

In contrast, belief contradictions (explicit disagreements about task
state or strategy) exhibited greater temporal volatility. Many teams
substantially reduced contradictions between Level 1 and Level 4,
indicating successful negotiation of shared interpretations through
dialogue and experience. However, some teams maintained or even
increased belief contradictions across levels, suggesting fundamental
misalignments that dialogue alone cannot resolve. The rarity of false beliefs and unsupported beliefs, combined with
their clustering in high-discrepancy teams, suggests these categories
may serve as markers of more fundamental coordination dysfunction
rather than routine coordination challenges. This asymmetric predictability suggests volatile teams may struggle with trust or strategic alignment, leading to improvements that fail to stabilize.
\subsection{Practical Implications}

First, the predictive
value of early-stage discrepancy patterns enables proactive
identification of at-risk teams. Rather than waiting for performance
failures to manifest, organizations could monitor dialogue-based
discrepancies during initial collaboration to flag teams likely to
experience future coordination breakdowns. Second, the distinct patterns across discrepancy types suggest that
interventions should be tailored to specific coordination
challenges. Teams struggling primarily with omissions might benefit
from structured communication protocols or explicit common ground
establishment procedures. Teams exhibiting persistent belief
contradictions might require facilitated discussion to surface and
resolve fundamental disagreements about task interpretation or
strategy.

\subsection{Limitations and Future Directions}

Some limitations merit consideration. Our task structure (asymmetric information object identification) may not generalize to symmetric tasks, distributed teams, or longer collaborations. The additive prediction model assumes independence across discrepancy types, which is unlikely — high omission rates likely cascade into more belief contradictions, and timing of discrepancies within levels is ignored entirely. Future work should incorporate team-specific learning curves, recency weighting, and type-specific dynamics to improve prediction accuracy.

\section{Acknowledgments}
This work was in part funded by DARPA contract \#HR001124C0502.
\printbibliography

\end{document}